# Aboveground biomass mapping in French Guiana by combining remote sensing, forest inventories and environmental data

Ibrahim Fayad[1], Nicolas Baghdadi[1], Stéphane Guitet[2], Jean-Stéphane Bailly[3], Bruno Hérault[4], Valéry Gond[5], Mahmoud El Hajj[6], Dinh Ho Tong Minh[1]

[1] IRSTEA, UMR TETIS, 500 rue Jean François Breton, 34093 Montpellier Cedex 5, France

[2] Office National des Forêts (ONF), R&D department, Cayenne, French Guiana,

[3] AgroParisTech, UMR LISAH, 2 place Pierre Viala, 34060 Montpellier, France

[4] INRA, UMR EcoFoG, BP 316 – 97379, Kourou cedex, French Guiana

[5] CIRAD, UPR B&SEF, campus de Baillarguet, 34398 Montpellier Cedex 5, France

[6] NOVELTIS, 153 rue du Lac, 31670 Labège, France

*Abstract*

**Mapping forest aboveground biomass (AGB) has become an important task, particularly for the reporting of carbon stocks and changes. AGB can be mapped using synthetic aperture radar data (SAR) or passive optical data. However, these data are insensitive to high AGB levels (>150 Mg/ha, and >300 Mg/ha for P-band), which are commonly found in tropical forests. Studies have mapped the rough variations in AGB by combining optical and environmental data at regional and global scales. Nevertheless, these maps cannot represent local variations in AGB in tropical forests. In this paper, we hypothesize that the problem of misrepresenting local variations in AGB and AGB estimation with good precision occurs because of both methodological limits (signal saturation or dilution bias) and a lack of adequate calibration data in this range of AGB values. We test this hypothesis by developing a calibrated regression model to predict variations in high AGB values (mean >300 Mg/ha) in French Guiana by a methodological approach for spatial extrapolation with data from the optical geoscience laser altimeter system (GLAS), forest inventories, radar, optics, and**

environmental variables for spatial inter- and extrapolation. Given their higher point count, GLAS data allow a wider coverage of AGB values. We find that the metrics from GLAS footprints are correlated with field AGB estimations ($R^2$=0.54, RMSE=48.3 Mg/ha) with no bias for high values. First, predictive models, including remote-sensing, environmental variables and spatial correlation functions, allow us to obtain "wall-to-wall" AGB maps over French Guiana with an RMSE for the *in situ* AGB estimates of ~51 Mg/ha and $R^2$=0.48 at a 1-km grid size. We conclude that a calibrated regression model based on GLAS with dependent environmental data can produce good AGB predictions even for high AGB values if the calibration data fit the AGB range. We also demonstrate that small temporal and spatial mismatches between field data and GLAS footprints are not a problem for regional and global calibrated regression models because field data aim to predict large and deep tendencies in AGB variations from environmental gradients and do not aim to represent high but stochastic and temporally limited variations from forest dynamics. Thus, we advocate including a greater variety of data, even if less precise and shifted, to better represent high AGB values in global models and to improve the fitting of these models for high values.

*Keywords: Aboveground biomass mapping; LiDAR; ICESat GLAS; forests; French Guiana*

## 1. Introduction

Mapping and characterizing spatial variations in forest aboveground biomass (AGB) are important tasks, particularly for the reporting of carbon stocks and changes. Recently, a considerable effort has been made to better quantify the amounts of AGB, especially in neo-tropical rain forests (e.g., Saatchi et al., 2011; Baccini et al., 2012; Mitchard et al., 2014; Avitabile et al., 2015). However, AGB, and thus forest carbon stocks, must be estimated with an acceptable precision to use these products in various forest management and monitoring projects, which still faces major challenges ( Mitchard et al., 2014).

Existing AGB estimation methods from satellite remote sensing data can be classified into three techniques: LiDAR, radar, and optical imagery. These methods are either limited to low AGB levels (<150 Mg/ha) (sensor saturation at certain biomass levels when using mainly radar and optical data) or have a limited spatial coverage (when using LiDAR data). Methods that use L-band radar and optical data to estimate the AGB are successful in forests with low to medium levels of AGB (e.g., Mitchard et al., 2012; Sandberg et al., 2011; Le Toan et al., 2011; Ploton et al., 2013; Lu et al., 2012). Indeed, current techniques that are based on passive optical sensing have shown limited sensitivity to biomass with the use of medium to high resolution imagery when the biomass reaches intermediate levels (150-200 Mg/ha) (e.g., Ploton et al., 2013; Lu et al., 2012). In contrast, very-high-resolution optical image textural analyses, e.g., Fourier Transform Textural Ordination (FOTO), have been used for non-saturating estimates of tropical forest biomass estimation. As such, this approach may provide higher sensitivity to biomass high levels (>600 Mg/ha) (e.g., Couteron et al., 2005; Proisy, Couteron and Fromard, 2007; Barbier et al., 2010). Another approach for AGB estimation over large scales is classifying forest covers by using optical remote sensing data and attributing a unique AGB value per forest type from field data averaging. Synthetic aperture radar (SAR) systems such as PALSAR/ALOS, JERS-1 and SIR-C, along with airborne SAR such as SETHI Bonin and Dreuillet, 2008 and E-SAR Horn, 1996, have also been used as an alternative for biomass estimation. However, these systems, as with passive optical sensing, also showed limited sensitivity to biomass depending on the characteristics of the forest, with maximum biomass values of 150 Mg/ha in the L-band (e.g., Baghdadi et al., 2014; Mitchard et al., 2012; Le Toan et al., 2011). Nonetheless, the use of SAR sensors with higher radar wavelengths (P-band, for example) may allow the estimation of biomass at higher biomass levels (on the order of 300 Mg/ha, Tong Minh et al., 2014).

LiDAR systems can capture the horizontal and vertical structure of vegetation in great detail; thus, LiDAR can estimate biomass with better precision in comparison to the techniques using radar and optical data ( Nelson et al., 2009; Dubayah et al., 2010). In addition, LiDAR has a much higher sensor saturation threshold to biomass (e.g., biomass estimation on the order of 1200 Mg/ha Lefsky et al., 2002). LiDAR systems acquire data either as a form of point clouds by using small footprints (<1 m) from airborne LiDAR systems or as waveforms by using large footprints (>10 m) from either airborne or spaceborne LiDAR systems ( Blair, Rabine and Hofton, 1999; Abshire et al., 2005). Estimating biomass with airborne LiDAR data is more accurate ( Ni-Meister et al., 2010; Zolkos, Goetz and Dubayah, 2013) but is generally limited to small areas given the high acquisition costs ( Avitabile et al., 2012). GLAS LiDAR data, on the other hand, are freely accessible and globally available but with sparse, inhomogeneous and anisotropic coverage.

Finally, field AGB estimates are provided by forest inventory plots. These estimations usually require measuring tree characteristics within a plot (always diameter at breast height, often tree height, and sometimes local wood densities). The AGB for each tree is then estimated by using allometric equations ( Chave et al., 2005). Previous studies developed plot AGB estimation techniques to estimate the AGB at the plot level. Despite some irreducible error sources, these methods provide the most accurate AGB estimations (Molto, Rossi and Blanc, 2013). However, over large areas, especially at a country level or larger, this method is hindered by several constraints, including a lack of field data in remote areas, inconsistent data collection methods over large regions, resources and time ( Boudreau et al., 2008). Therefore, remote sensing data are used to extrapolate AGB estimates from the plot level to larger areas. However, no remote sensing instrument that can provide direct measurements of biomass has been developed, so field-sampled biomass is required to establish relationships between remote sensing signals and biomass to estimate the AGB at large scales ( Rosenqvist et al., 2003).

Given the complementary limitations of each system (AGB saturation of passive optical and radar systems and sparse coverage of LiDAR systems), fusing data from these systems is crucial to develop biomass maps over areas at the country or global levels. Recently, studies have used optical imagery, radar and LiDAR data to create AGB maps over large areas alongside optical and radar imagery to provide information on the land cover, vegetation types, forest coverage, and LiDAR data to capture the vertical structure of forests ( Sun et al., 2011; Koch, 2010).

Boudreau et al., 2008 first estimated the biomass in Quebec, Canada by using GLAS waveform metrics, a Landsat ETM+ land cover map, a Shuttle Radar Topographic Mission (SRTM) digital elevation model, ground inventory plots, and vegetation zone maps. An allometric equation was initially used to predict the tree biomass by using tree attributes. Next, a generic airborne LiDAR-based biomass equation ($R^2$=0.65) was developed and used to estimate the biomass from LiDAR data at the GLAS footprint scale Boudreau et al., 2008. Thus, GLAS data allowed AGB estimates to be extrapolated and map the forest biomass along forest types that were derived from Landsat-7 ETM+ ( Wulder et al., 2012). The results showed an RMSE for the AGB estimates of 39.9 Mg/ha. Mitchard et al., 2012 estimated the AGB in Lopé National Park in central Gabon by using a combination of GLAS data and L-band PALSAR data. Because GLAS waveforms can estimate Lorey's height with confidence in tropical dense rainforests (Lefsky, 2010), these authors first found a high correlation between Lorey's height and the field AGB. This correlation allowed them to add 7042 GLAS-based AGB estimates to their field estimate. Finally, radar images and DEM data were jointly used to classify forest vegetation into a homogeneous set, enabling the spatial extrapolation of AGB estimates. The results showed that AGB estimates at the plot level have a ±25% uncertainty (%RMSE). In addition to regional AGB estimates, several attempts have been made to estimate the AGB at the pan-tropical scale (e.g., Saatchi et al., 2011; Baccini et al., 2012; Avitabile et al., 2015). Baccini et al., 2012 estimated the carbon density for the pan-tropics at a grid size of 500 m from MODIS data by using GLAS waveform metrics. Saatchi et al., 2011 mapped the biomass in tropical regions across three continents by using a combination of data from 4079 in situ inventory plots and GLAS samples of forest structure in addition to optical and microwave imagery (1-km resolution). These authors developed continent-based allometric equations to provide the best models to convert Lorey's heights from GLAS data to AGB. A data fusion model that was based on the maximum entropy (MaxEnt) approach was used to extrapolate the AGB from GLAS footprints to the entire landscape. Despite the progresses in terms of biomass mapping over large scales, the reliability of most of these global products has proven to be low. Indeed, a comparison by Mitchard et al., 2013 of the global AGB maps in Saatchi et al., 2011 and Baccini et al., 2012 showed that both products showed large discrepancies over smaller spatial scales, even though they might agree on the country scale (Mitchard et al., 2013). Both maps showed high bias with very large uncertainties (RMSE >60 Mg/ha) compared to a more precise AGB map of Amazonian Columbia (Mitchard et al., 2013). Moreover, both maps had different spatial patterns of AGB with little consistency, even though both studies used the same methodologies, same primary data source (GLAS LiDAR) and a similar modelling approach for AGB mapping. Overcoming this bias problem is an important challenge to generate precise estimates of the carbon flux in tropical forests, particularly in the Guiana shield, where the AGB stocks are very high and where global maps show the largest errors, which exceed 60 Mg/ha (Avitabile et al., 2015). We hypothesize that one of the reasons for the presence of biases and inconsistencies in these products originate from both saturation and a lack of prediction data during model calibration in high biomass areas, which limits the efficiency of the models in these ranges. In fact, AGB plots are much more difficult and expensive to implement in rainforests than in other and more accessible types of vegetation. Thus, models that are fitted on many low AGB plots and few high AGB plots correctly report the contrasting AGB between vegetation types (bushes, dry forests, moist forests, etc.) but fail to represent the variability of AGB within rainforests, which is much more important for carbon balance issues. We hypothesize that the same methods, which use GLAS LiDAR combined with imagery and radar products, could produce more efficient models by expanding the source of the field data and focusing on high biomass areas to calibrate the models.

We propose a calibrated regression model to map the AGB with different grid sizes (500, 1000, and 2000 m) by using LiDAR, SAR, optical and environmental data. Our aims are to (i) decrease the bias and increase the precision of the AGB estimates (which can be linked to the saturation of the radar and optical data) using GLAS data. GLAS data produce a better vertical presentation of forest structures without signal saturation. (ii) Capture better local AGB variations and patterns by combining LiDAR, radar, optical, environmental data and as much field data as possible. (iii) Finding the optimal scale that allows averaging the dynamic effects while capturing maximum environmental and spatial effects. A description of the different datasets used in this study is provided in Section 2. Section 3 presents the methodology for the creation of a wall-to-wall AGB map. The results are shown in Section 4. Finally, Sections 5 and 6 present the discussion and conclusion sections, respectively.

## 2. Study area and datasets

### 2.1. Study area

French Guiana is located on the northern coast of the South American continent and borders the Atlantic Ocean, Brazil and Suriname. The study site's area is approximately 83500 km$^2$, where the forest covers more than 80000 km$^2$. The terrain is mostly low-lying (70% of slopes are less than 5°), rising occasionally to small hills and mountains, with an altitude that ranges from 0 to 851 m ( Guitet et al., 2013). French Guiana has an equatorial climate with two main seasons: a dry season from August to December and a rainy or wet season from December to June.

### 2.2. Datasets

#### 2.2.1. GLAS dataset

LiDAR data were acquired from GLAS on board the Ice, Cloud, and Land Elevation Satellite (ICESat) between 2003 and 2009. The GLAS laser footprints have a nearly circular shape of approximately 70 m in diameter and a footprint spacing of approximately 170 m along their track.

The data were acquired during 18 missions by using three on-board lasers with orbit cycles that repeated between 57 and 197 days.

The horizontal geolocation error of the ground footprints is less than 5 m on average for all ICESat missions (http://nsidc.org/data/icesat/laser_op_periods.html). Several studies (e.g., Carabajal and Harding, 2006; Huang et al., 2011) have estimated the vertical accuracy of the GLAS to be between 0 and 3.2 cm on average over flat surfaces.

The GLA01 and GLA14 data products that were available for ICESat/GLAS were used in this study. GLA01 comprises the full waveform data, and GLA14 comprises the global land surface altimetry data. Over flat terrain, the waveforms that were acquired over vegetated areas are bimodal distributions and represent the different layers (top of canopy, branches, ground, etc.) of the GLAS footprint. Several filters were applied to exclude unreliable GLAS data (i.e., data that were affected by atmospheric conditions, clouds, etc.):

(1) Signals with high noise were removed when the signal-to-noise ratio was higher than 15 (e.g., Carabajal and Harding, 2006; Chen, 2010; Lee et al., 2011).

(2) GLAS waveforms with delays from either saturation or atmospheric forward scattering were removed. Only cloudless waveforms were kept by using the cloud detection flag (FRir_qaFlag = 15). Saturated signals were identified by using the GLAS flag (SatNdx >0). Both the FRir_qaFlag and SatNdx flags are found in the GLA14 product.

(3) Waveforms with a centroid elevation that were significantly higher or lower than the corresponding SRTM elevation were removed (|SRTM - GLAS| >100 m) ( Baghdadi et al., 2014).

From the original database of 101312 footprints, 39856 footprints that satisfied the three filter conditions were kept.

Some metrics were derived from the GLAS waveforms to reflect the canopy's vertical variables. First, the waveform extent (Figure 1, Wext), which is defined as the difference between the signal's end and beginning, was calculated for each GLAS waveform. A signal's beginning and end (Figure 1) are defined as the first and last locations where the waveform intensity exceeds a noise threshold that equals 4.5 times the standard deviation of the background noise (Lefsky et al., 2007).

In addition, Gaussian peaks from the decomposition of each GLAS waveform, which represent canopy features such as the canopy top (position of the first Gaussian peak, Figure 1), canopy trunks, ground or a mix of these elements, were identified. However, the last Gaussian peak does not necessarily represent the ground return (e.g., Chen, 2010; Rosette, North and Suarez, 2008; Duong et al., 2009; Sun et al., 2008). Duong et al., 2009 and Sun et al., 2008 identified the ground as the last peak. Rosette, North and Suarez, 2008 and Chen, 2010 found that the elevation of the stronger of the last two Gaussian peaks has a better correspondence to the ground. In this study, the stronger of the last two Gaussian peaks was selected as the ground peak (Figure 1). After identifying the canopy top and ground peaks, the quartile heights of the GLAS waveforms (10 through 90%) were calculated from the signal's beginning.

In addition, the leading edge (Lead), which is defined as the elevation difference between the signal's beginning and the canopy top, and the trailing edge (Trail), which is defined as the difference between the signal's end and the ground peak's center (Hilbert and Schmullius, 2012), were calculated (Figure 1).

**Figure 1. A typical GLAS waveform acquired over a vegetated area on flat terrain.**

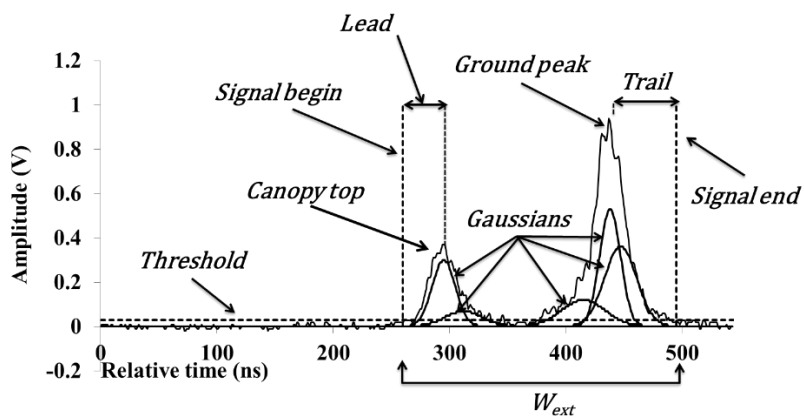

### 2.2.2. MODerate-resolution Imaging Spectroradiometer (MODIS) data

The MODIS sensors mounted on the Terra and Aqua satellites possess a total of 36 spectral bands, seven of which are designed specifically for land applications with spatial resolutions that range from 250 m to 1 km. The MODIS dataset (MOD13A1c5) that was used in this study includes ten years (January $1^{st}$, 2003 to December $31^{st}$, 2012) of enhanced vegetation index (EVI) time series data with a resolution of 250 m. These EVI data effectively characterize biophysical and biochemical states and processes from vegetated surfaces. Freitas, Mello and Cruz, 2005 and Pascual et al., 2010 found a strong relationship between forest structures (forest height and biomass) and vegetation indices. This 10-year period was synchronized with the GLAS data (from 2003 to 2009). Several maps were created using these EVI time series data: (1) minimum, mean and maximum values of the EVI time series data (MIN_EVI, MEAN_EVI, and MAX_EVI respectively); (2) the first seven principal components that were issued from the principal component analysis of the EVI time series data (PC1, PC2 … PC7); and (3) eight statistical maps ("mean", "variance", "homogeneity", "contrast", "dissimilarity", "entropy", "second_moment", and "correlation") that were derived from the Gray-Level Co-Occurrence Matrix (GLCM) method by using the mean EVI map and a 3x3 window size. GLCM functions characterize the texture of an image by calculating how often pairs of pixels with specific values and in a specified spatial relationship occur in an image. After creating a GLCM, the aforementioned statistical measures can be calculated from this matrix. Figure 2 shows the map of the mean EVI time series data.

**Figure 2. Mean values of the MODIS EVI time series data (January 1$^{st}$, 2003 to December 31$^{st}$, 2012).**

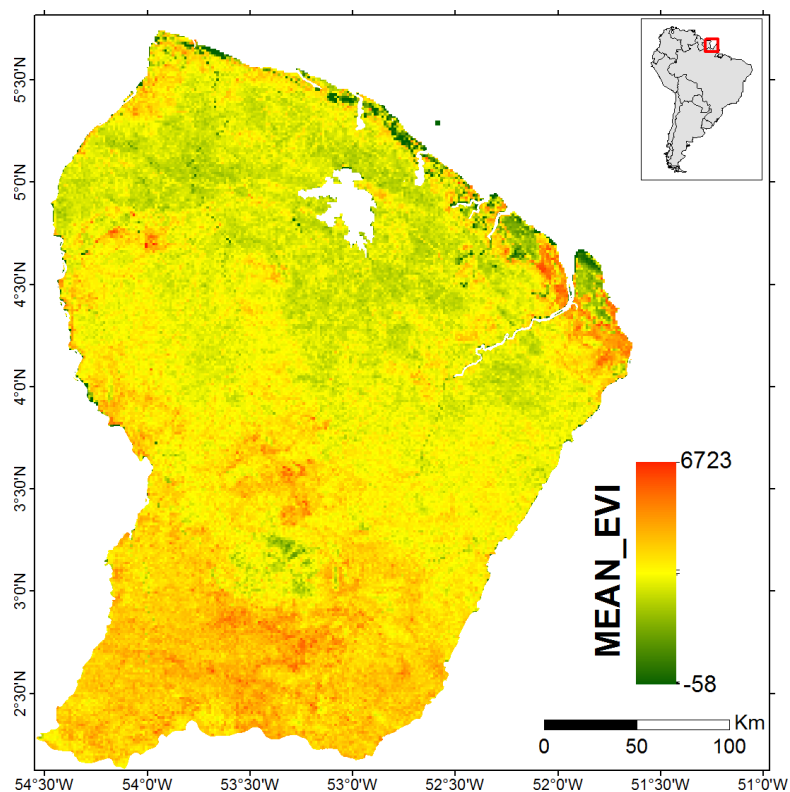

### 2.2.3. SRTM digital elevation model data

The Shuttle Radar Topography Mission (SRTM) digital elevation model, which has a resolution of 90 m, was used in this study. This SRTM DEM dataset and its derived maps were considered because the local topography and drainage are important for tree anchorage and forest dynamics both directly and through soil types. Three derivative maps were created from these SRTM DEM data: (1) a slope map (Slope); (2) a surface roughness map (Rug), which was defined as the standard deviation of the elevation in a 3x3 moving window; and (3) a map of the heights above the nearest drainage (HAND). The HAND normalizes the topography according to the local relative heights along the drainage network and thus presents the topology of the relative soil gravitational potentials or local draining potentials (Nobre et al., 2011).

### 2.2.4. Geological map

The geology is an important determinant of soil formation and the conditioning of the chemical and physical properties in the soil, which affects tree growth and other forest parameters. A geological substratum map (GEOL) that was produced by the French Geological Survey (Delor et al., 2001) was used in this study (Figure 3a). The map was simplified to retain only five main rock formations: recent sediments, volcanic sedimentary rock, granites, gabbros, and gneiss.

### 2.2.5. Forest landscape type map

A forest landscape type map of the Guiana shield that was developed by Gond et al., 2011 at 1-km resolution was also used (Figure 3b). In this map, 33 remotely sensed landscape types (LTs) were interpreted by using VEGETATION/SPOT images. Over French Guiana, the forested landscapes consist of only five classes out of these 33 (with less than five percent of water bodies and non-forested areas): a low dense forest (LT8), a high forest with a regular canopy (LT9), a high forest with a disrupted canopy (LT10), a mixed high and open forest (LT11), and an open and wet forest (LT12).

### 2.2.6. Forest landscape level map

A forest landscape level map (LLM) that covers French Guiana, was also used in this study. The LLM was created in a study by Guitet et al., 2013. The map was produced by first segmenting French Guiana's terrain into 224000 landform units by using a modified counting box algorithm.

Next, Principal Components Analysis (PCA) followed by k-means clustering (Ward's method) identified 12 different landform types that corresponded to theoretical elementary landforms. The forest landscape-level map variable has shown a very strong relationship with the biomass according to a study by Guitet et al., 2015.

### 2.2.7. Average rainfall map

In addition, precipitation data from NASA's tropical rainfall measuring mission (TRMM) were used. The TRMM data that were used in this study correspond to the average daily precipitation over the last 10 years (2003-2013) with a resolution of 8 km (Rain). Rainfall was used in AGB mapping because of the strong relationship between precipitation and biomass (e.g., Silvertown et al., 1994, Hong et al., 2015).

**Figure 3. Geological map (a) and forest landscape type map (b).**

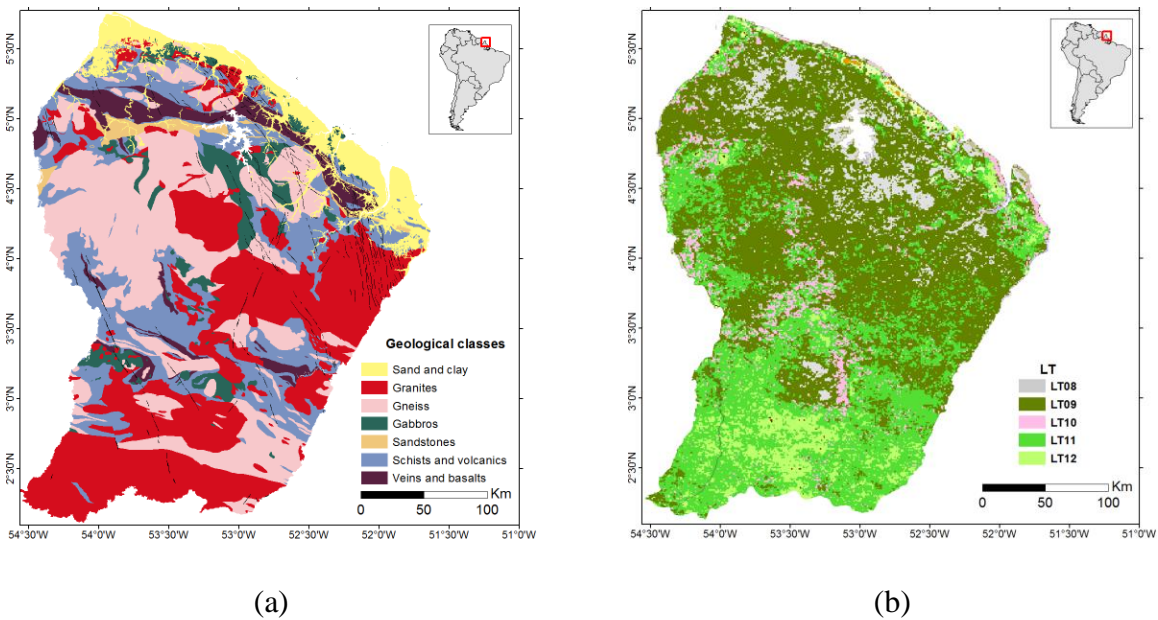

(a)                                                                 (b)

### 2.2.8. L-band PALSAR images

A 25-m mosaic of L-band high-penetration depth images from the PALSAR radar, which were acquired in 2010 over French Guiana, was used in this study. The mosaic is available in both HH and HV polarizations. However, the HV polarization is more sensitive to biomass than HH ( Tong Minh et al., 2014), so only the HV polarization was considered in this study. The original 25-m resolution map was resampled according to the desired AGB map's grid size (500, 1000 and 2000

m) by using either the mean or the median as the resampling scheme. The GLCM texture indices were calculated on 3x3 windows from the resampled images. The variable names were the texture indexes, which were preceded by "Rad_avg" and "Rad_med" for the texture indices from the radar images that were resampled by the mean and median, respectively.

### 2.2.9. Canopy height map

A canopy height map that was produced by Fayad et al., 2016 was also used in this study (Figure 4). The map was created by a data fusion of airborne LiDAR data (covering 4/5 of French Guiana with a point density of 1.59 pts/km$^2$), environmental and optical data. To create this map, the random forests (RF) algorithm was used to find a relationship between the airborne LiDAR canopy heights and the auxiliary data, which allowed us to estimate the canopy heights over all of French Guiana with a grid size of 250 m. Next, adding the kriged model residuals to the map increased its precision to 3.3 m (Figure 4). Kriged model residuals refer to the ordinary kriging of regression (RF) residuals that are assumed to present order-2 stationary spatial covariance. Several maps were produced and used to create height maps with grid sizes of 500, 1000, and 2000 m: (1) the height mean (H_x_Avg) and height median (H_x_Med) values for 2x2, 4x4 and 8x8 cell sizes; (2) the mean value (H_x_LoMean) of the 2, 4, and 8 smallest canopy heights for 2x2, 4x4 and 8x8 cell sizes; and (3) the mean value of the 4 highest canopy heights (H_x_HiMean) for 4x4 and 8x8 cell sizes.

**Figure 4. Canopy height map of French Guiana with a grid size of 250 m.**

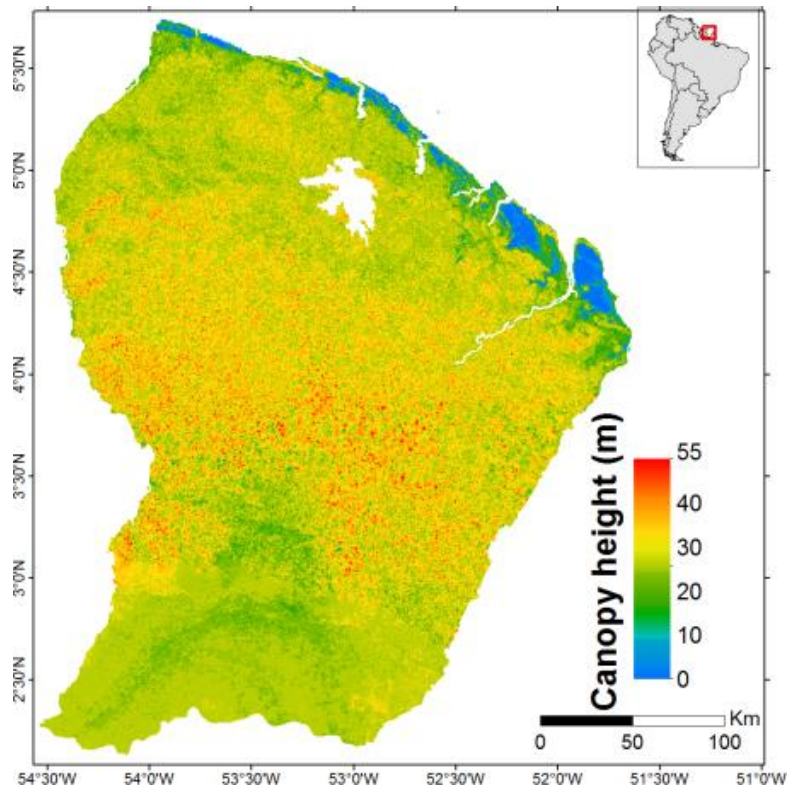

### 2.2.10. *In situ* AGB dataset

Two different forest inventories were used as field biomass estimates in this study ( Guitet et al., 2015). The first dataset was acquired by the CTFT ("Centre Technique Forestier Tropical") between 1974 and 1976 for 14 blocks, which covered 547 000 ha in the northern part of the French Guiana (FG) territory (Baccini et al., 2008). To remove uncertainties from the CTFT dataset, we excluded plots with partial inventories (pre-census DBH equal to 40 cm) or containing several forest types (i.e. mixed terra firme and seasonally flooded forest). Plots located in areas affected by forest harvesting or gold mining between 1974 and 2007 were also eliminated. The second inventory was campaigned between 2006 and 2013 on behalf of the ONF ("Office National des Forêts": the French national forest agency) to complete the regional coverage and better sample the environmental variability (Clark et al., 2011). The two forest inventories covered a total of 1120 ha and inventoried on 2507 plots, equaling an overall sampling rate of 0.014 % when converted to the total French Guiana area, with local sampling rates that varied between 0.2 and 1%. The biomass was estimated by using the generic pan-tropical allometry (equation 1) from

Chave et al., 2014 and Monte-Carlo simulations to consider the incertitude of forestry data (refer to Guitet et al., 2015 for more details).

$$AGB_{tree} = 0.0673 \times (WSG \times DBH^2 \times H)^{0.973} \tag{1}$$

where $AGB_{tree}$ is the AGB (g) at the tree scale, WSG (g/cm$^3$) is the tree wood density, DBH (m) is the diameter breast height and H (m) is the tree height.

The distribution of AGB estimates between the two campaigns doesn't show any significant differences despite the long-time interval between the two campaigns and the difference in field methods (Kolmogorov-Smirnov test, D = 0.0299, p-value = 0.631). At plot scale, CV rarely exceeded 17% for CTFT data (10% for ONF). The confidence interval of mean AGB estimates at 95% at the plot scale did not exceed 6 Mg/ha at the worst case. Moreover, the monitored permanent plots in French Guiana (on Paracou for the last 30 years) showed no significant trend in the evolution of biomass (increase in general but not significant if one takes into account the few events of death). Therefore, we can hypothesize that the old inventory reflects correctly the mean variations of biomass linked to environmental and spatial effects.

The *in situ* AGB dataset was randomly divided into two datasets with 1253 plots each. One dataset was used for model building and the other to validate the wall-to-wall AGB maps.

## 3. Materials and methods

Our methodology to map the aboveground biomass (AGB) consisted of four steps of spatial modelling within the Kriging regression framework (Hengl, Heuvelink and Stein, 2004). (i) Finding the best regression between GLAS waveform-derived metrics and *in situ* AGB, with additional GLAS-derived estimates used as additional AGB data. (ii) Mapping AGB at several grid sizes (500 m, 1000 m, and 2000 m) by developing a trend model of AGB data from GLAS estimates with predictors that are derived from optical, radar, and environmental maps. (iii) Improving the AGB map's precision by adding the kriged residuals to the AGB map from step (ii). (iv) Finally, estimating the map's efficiency by using the validation dataset.

## 3.1. AGB estimation of GLAS waveforms

We used a linear model (LM) of all the GLAS metrics in section 2.2.1 to quantify to what extent AGB can be described with these GLAS metrics. Stepwise regression based on the Bayesian information criterion (BIC) was conducted to identify the most predictive GLAS metrics for the linear regression model to determine the GLAS metrics that had high correlations with the biomass and to discard the metrics that had less effect on the AGB estimation. The coefficient of determination ($R^2$), which is widely used to evaluate the goodness of fit for regression models, was used to analyze the performance of the models. The root mean square error (RMSE) was also computed to assess the predicted AGB versus AGB that was estimated from field observations at the footprint scale. Both the $R^2$ and RMSE are obtained from the k-fold cross validation of the models.

The GLAS footprints did not exactly intersect any inventory plots that were acquired over French Guiana, so we chose a maximum distance between the *in situ* AGB estimates and the GLAS footprints to relay the GLAS metrics to the AGB. The distance was selected based on the performance of a linear regression model for AGB estimation using the GLAS metrics. Each dataset of this model represented GLAS data from a different maximum distance from the *in situ* data. The selected maximum distance must maximize the number of available *in situ* AGB estimates while retaining the highest possible coefficient of determination. Six maximum distances were tested: 100, 200, 250, 300, 350, and 400 m. Choosing a distance of 50 m was not possible because of the very low number of available in situ data (9 AGB plot estimates). The results showed that the linear regression model produced an $R^2$ of 0.63 when using a distance of 100 m, with only 21 *in situ* AGB estimate plots available. This $R^2$ decreased to 0.56 (41 plots), 0.53 (89 plots), 0.39 (166 plots), 0.37 (281 plots), and 0.11 (374 plots) for distances of 200, 250, 300, 350, and 400 m, respectively (Figure 5). Hence, a distance of 250 m was chosen as the best trade-off between an $R^2$ of 0.53 and a number of points of 89. Finally, the linear regression model was applied to the remainder of the GLAS footprints (these 39825 GLAS AGB estimates will be used later as calibration data in the wall-to-wall mapping algorithm).

**Figure 5. Effects of the distance between the GLAS footprints and *in situ* AGB estimate plots on the correlation between the GLAS metrics and *in situ* AGB estimates.**

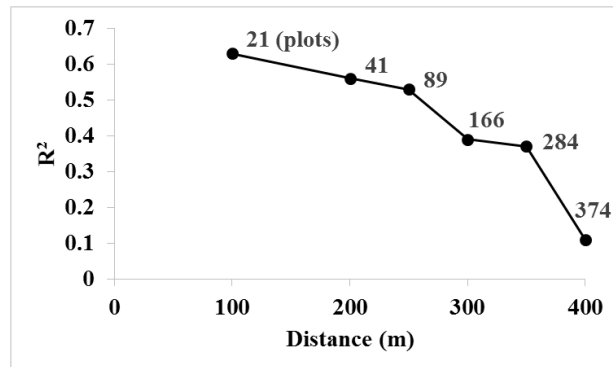

## 3.2. Mapping wall-to-wall AGB

We used the ~40000 GLAS AGB estimates ($AGB_{GLAS}$) from section 3.1 as AGB calibration data to test two models for AGB wall-to-wall mapping over all of French Guiana: one with the linear regression model (LM) and the other with the random forests (RF) algorithm. The random forests (RF) technique is a performance regression method that is becoming widely used by the remote sensing community for biomass estimation and other studies (e.g., Baccini et al., 2012, Baghdadi et al., 2014). The main advantage of random forests is its incorporation of continuous or qualitative predictors without making assumptions regarding their statistical distribution or covariance structure ( Breiman, 2001). First, all the auxiliary variables were used in the LM and RF models for AGB estimation. Next, we used stepwise regression that was based on BIC for the LM model and the variable importance test for the RF model, and the best variables that explained the AGB were selected. The best model between LM and RF was chosen by using the highest $R^2$ and lowest RMSE. Finally, the independence of the residuals of the selected model was tested (Figure 6). Because the RF model failed to guarantee the spatial independence of the residuals, a spatial component (residual Kriging) was added to the AGB estimates map. This technique, which combines a regression model and Kriging, is known as regression Kriging (RK) (Figure 6). RK was developed primarily to consider the correlation between environmental variables and the unsatisfactory goodness of fit of the spatial variance model of the dataset (Sun, Minasny and McBratney, 2012). RK has the following form:

$$\hat{z}(s_0) = \hat{m}(s_0) + \hat{e}(s_0) = \hat{m}(s_0) + \sum_{i=1}^{n} \lambda_i . e(s_i) \qquad (2)$$

where $\hat{z}(s_0)$ is the predicted value at an unvisited location $s_o$, $\hat{m}(s_0)$ is the fitted trend (using LM or RF on environmental variables), and $\hat{e}(s_0)$ is the kriged residual. $\lambda_i$ are the Kriging weights, which are determined by the spatial autocorrelation structure (variogram) of the residual and the spatial data sampling scheme, and $e(s_i)$ is the residual at location $s_i$.

In this study, we used the ordinary Kriging (OK) model, which allows the interpolation of un-sampled data that are based solely on a linear model of regionalization, which is known as a semivariogram. This semivariogram plots the semivariance γ as a function of the distance between samples *h* by using the following function:

$$\gamma(h) = \frac{1}{2N(h)} \sum_{i=1}^{N(h)} [e(s_i) - e(s_i + h)]^2 \qquad (3)$$

where $\gamma(h)$ is the semivariance as a function of the lag distance *h*, $N(h)$ is the number of pairs of data that are separated by *h*, and e is the AGB estimate residuals from regression at locations $s_i$ and ($s_i$+*h*) ( Goovaerts, 1997). Semivariograms have three main parameters: (1) the *nugget*, which is the semivariance at a lag distance close to zero; (2) the *sill*, which is the semivariance where no spatial correlation exists at long distances; and (3) the *range*, which is the distance at which the *sill* is reached. After plotting the empirical semivariogram, which describes the spatial autocorrelation of a given dataset, a mathematical function is fitted to this empirical semivariogram to represent the spatial structure of the residuals. Thus, the dataset's empirical variogram can now be represented by using a function. Ordinary Kriging is used after the model fitting of the sample semivariogram, which estimates the values of ê at location $s_0$ by using the following equation:

$$\hat{e}(s_0) = \sum_{i=1}^{n} \lambda_i e(s_i) \qquad (4)$$

where $\hat{e}(s_0)$ is the predicted value at an unvisited location $s_0$ (in this case, the GLAS AGB estimate residual at location $s_0$), and $\lambda_i$ are the weights of n neighboring samples ( Goovaerts, 1997). The weights $\lambda_i$ depend on the fitted semivariogram function, the distance to the predicted location, and the spatial relationships among the measured values around the prediction location.

**Figure 6. AGB mapping procedure.**

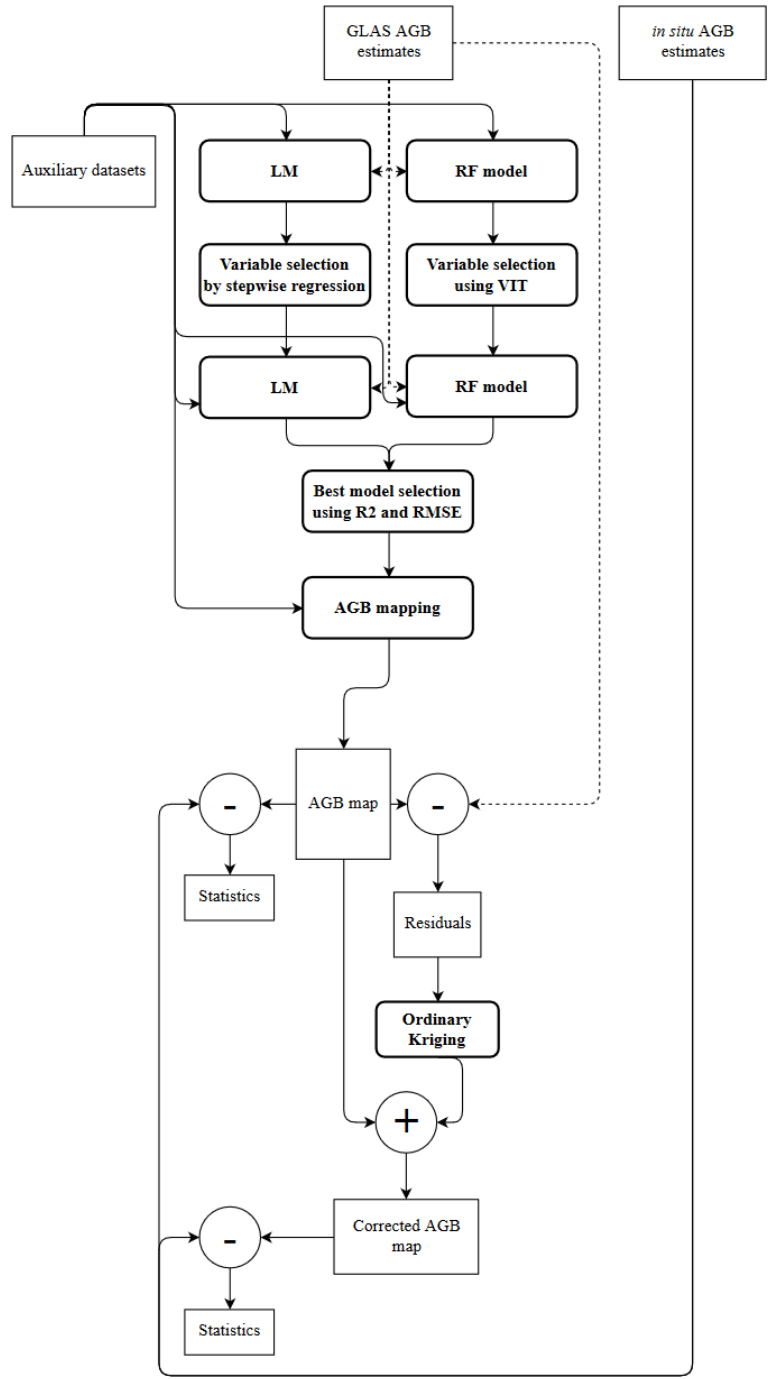

The wall-to-wall AGB maps were produced at three different grid sizes (500, 1000, and 2000 m). The auxiliary variables for each map were resampled (averaging the value of the cells) to the corresponding grid size. To compare the precision of each produced AGB map, the *in situ* AGB estimates were overlaid onto the different maps, and each cell that featured more than four *in situ* AGB estimates was selected. Next, the RMSEP (Root Mean Square Error of Prediction) and $R^2$ were calculated by using the mean values of the *in situ* AGB estimates.

## 4. Results
### 4.1. Relationship between GLAS metrics and AGB

The 89 *in situ* AGB estimate plots that neighbored the GLAS footprints at a distance of 250 m were used to create a model that linked the *in situ* AGB estimates and all the metrics that were extracted from GLAS and a DEM (Wext, H10 through H90 with a 10% step, TI, Slope, and Top of Canopy Heights TCH). The results showed that linking the GLAS and the DEM metrics to the biomass with the following equation ($AGB_{insitu}$ = Wext + H10 + H20 + H30 + H40 + H50 + H60 + H70 + H80 + H90 + TI + TCH + Slope) created an RMSE for the AGB estimates of 58.6 Mg/ha ($R^2$=0.41) (Figure 7a). The best variables (with stepwise regression by using BIC) to be used for AGB estimation with the linear regression approach are the Wext, TCH, Lead, TI, H20, and H80. Thus, the AGB estimation model that uses GLAS metrics has the following form:

AGB = Wext + Lead + TI + TCH + H20 + H80         (5)

Using the best GLAS metrics for AGB estimation produced better AGB estimates compared to the model that used all the metrics. Figure 7b shows that the AGB estimates were more precise, with an RMSE of 48.3 Mg/ha, when using equation 5 ($R^2$=0.54).

**Figure 7.** Comparison between *in situ* AGB estimates and AGB estimates from the linear regression model when using all the GLAS and DEM metrics (a) and the most important GLAS and DEM metrics (b).

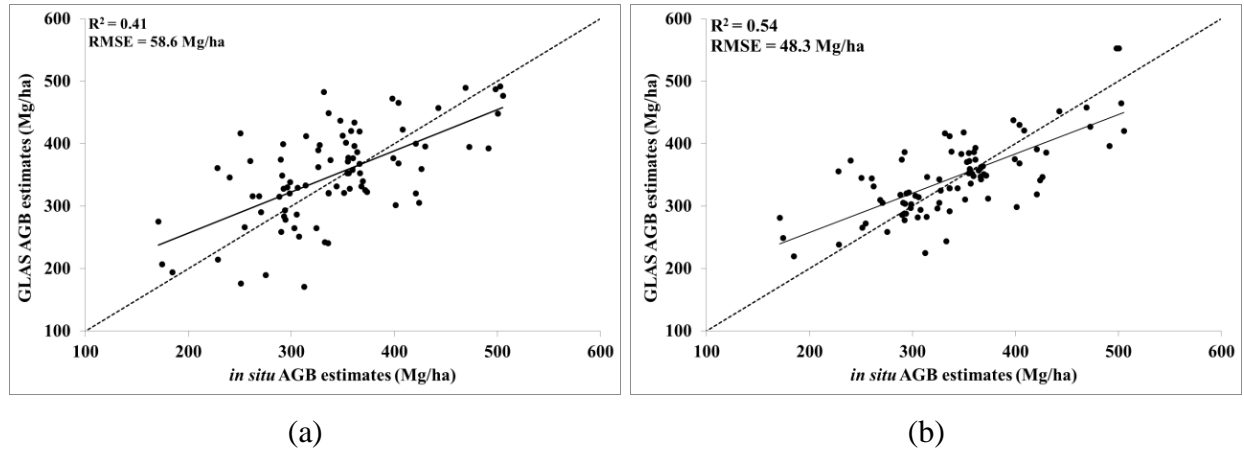

(a)   (b)

### 4.2. Mapping the AGB over French Guiana

To map the AGB over all of French Guiana, the linear regression model (equation 5) that was used to estimate the AGB from the GLAS metrics was applied to the remaining 39825 GLAS footprints (39825 footprints, excluding the 89 footprints that were used in section 4.1). These AGB estimates ensure larger coverage compared to only using the *in situ* AGB estimates. Next, two regression models (one linear and one that uses random forests) were created to link the 39825 GLAS AGB estimates ($AGB_{GLAS}$) to the auxiliary variables in section 2.2 at a 1000-m grid size. At this stage, we only chose the best model between the LM and RF for the wall-to-wall AGB mapping. Therefore, this test was only performed for the 1000-m grid size auxiliary data. First, the auxiliary variables that contributed the most to the estimation of the AGB for each model were determined. This process was conducted by using either Fisher's test (p-value) for the linear regression model or the increase in the mean square error of the predictions (%IncMSE, estimated with out-of-bag cross validation) from the variable importance test for the random forests model (average and standard deviations of 50 repetitions) (Figure 8). Model results showed that the precision of the modelled AGB estimates from the LM were 10% lower in comparison to the modelled AGB estimates from the RF model. Since the RF model performed better, it was used to obtain an initial

1000-m AGB map. The same procedure for variable selection, model calibration and AGB mapping was used to create two additional AGB maps at 500- and 2000-m grid sizes.

**Figure 8. Variables' order of importance in the AGB estimation models with the %IncMSE for the random forest regressions (left column, higher values are better) and 1 – p_value (right column, higher values are better).**

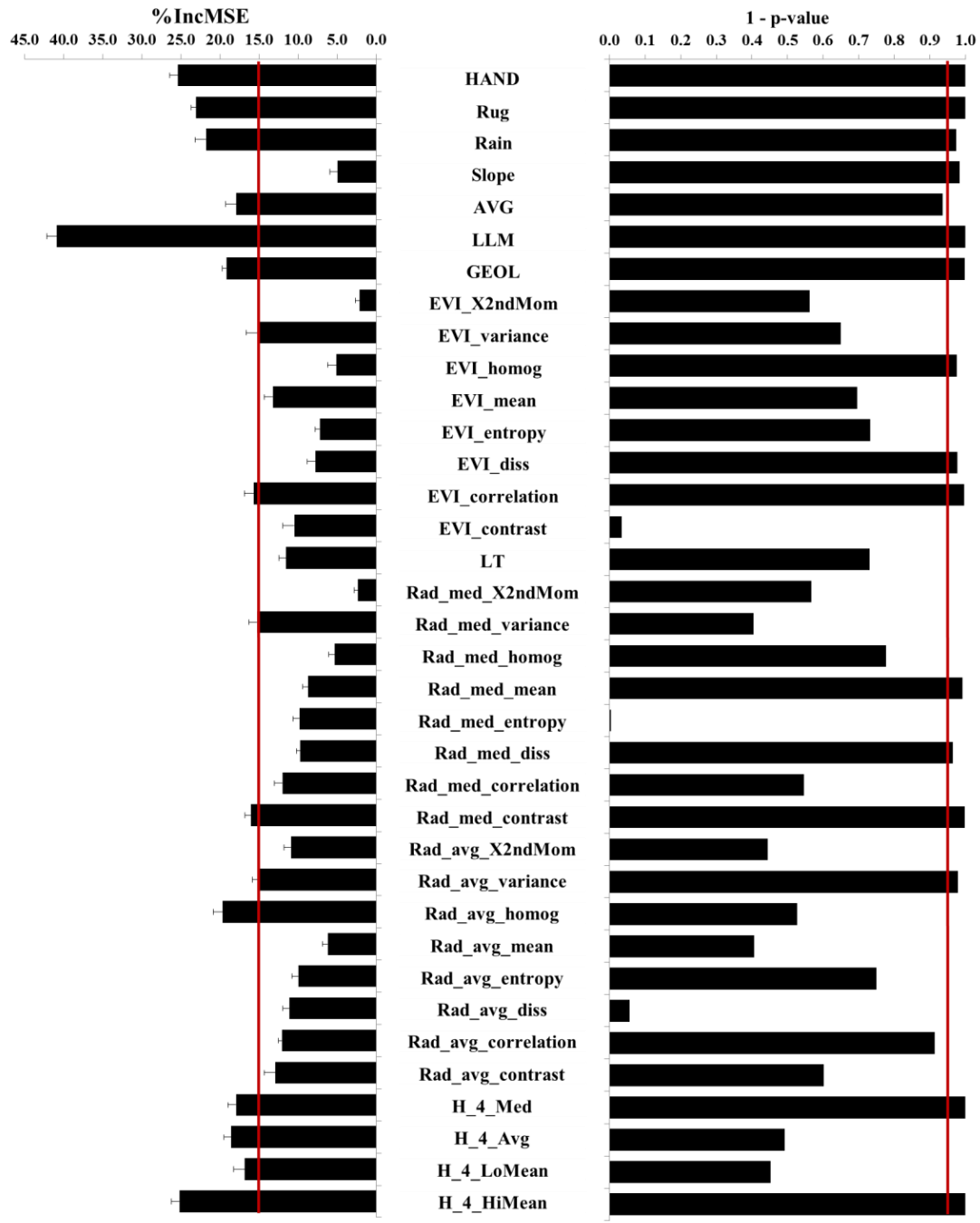

The precision of the three maps was compared to the *in situ* AGB estimates validation dataset (half the original *in situ* AGB estimates dataset). For each AGB pixel (from the 500-, 1000-, and 2000-m AGB maps), the precision was only calculated if that pixel contained at least four *in situ* AGB estimates (89, 143, and 195 validation pixels were found for the 500-, 1000-, and 2000-m AGB maps, respectively). The contained *in situ* AGB estimates were then averaged for that pixel. The results indicated that the precision of the created AGB maps was 73.1 Mg/ha ($R^2$=0.07), 53.4 Mg/ha ($R^2$=0.46), and 46.7 Mg/ha ($R^2$=0.48) for the 500-, 1000-, and 2000-m AGB maps, respectively (Figures 9a, 9b and 9c).

The regression Kriging technique was used to improve the precision of the AGB maps. This technique first required the subtraction of the mapped AGB from the GLAS AGB estimates. Next, the variogram of the residuals was calculated and represented with an exponential equation. The exponential equation for the 1000-m AGB map was presented by using a range of 3123 m, a sill of about 5500 and a nugget of about 9700. Furthermore, the residuals were kriged using the exponential equation, and the Kriging of the residuals was added to the original AGB map. The resulting map is presented in Figure 10. A comparison of the new AGB maps with the *in situ* AGB estimates showed an increase in the precision of the new maps, with an RMSE of 72.8 Mg/ha ($R^2$=0.12), 50.2 Mg/ha ($R^2$=0.66), and 43.2 Mg/ha ($R^2$=0.63) for the 500-, 1000-, and 2000-m AGB maps, respectively (Figures 9d, 9e and 9f).

**Figure 9. AGB estimation precision for three grid sizes, namely, 500 m (a and d), 1000 m (b and e), and 2000 m (c and f), before Kriging when using the RF technique (a, b, c) and after Kriging when using the regression Kriging technique (d, e, f).**

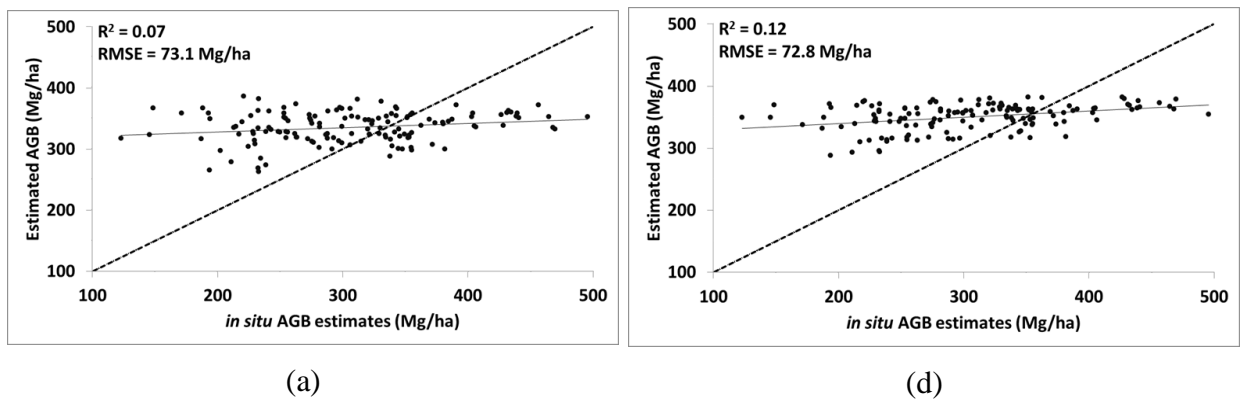

(a)                                                                    (d)

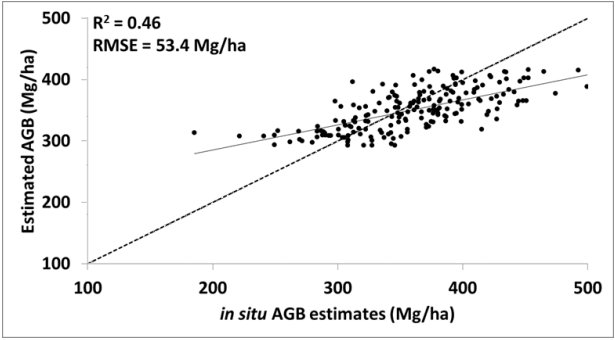

(b)

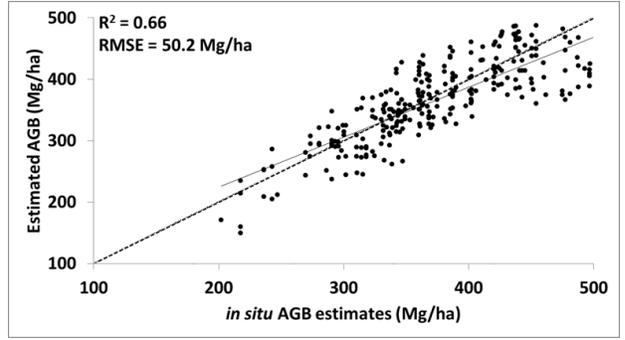

(e)

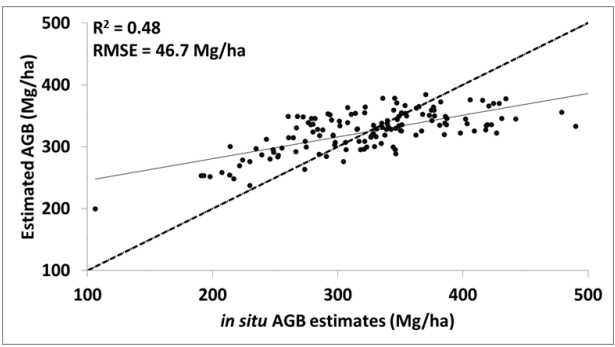

(c)

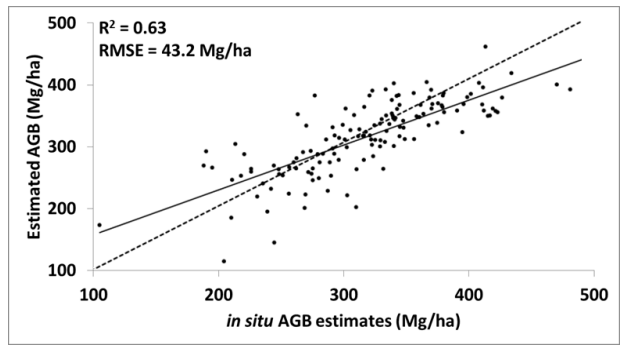

(f)

**Figure 10. A 1-km grid-size AGB map of French Guiana from the regression Kriging technique.**

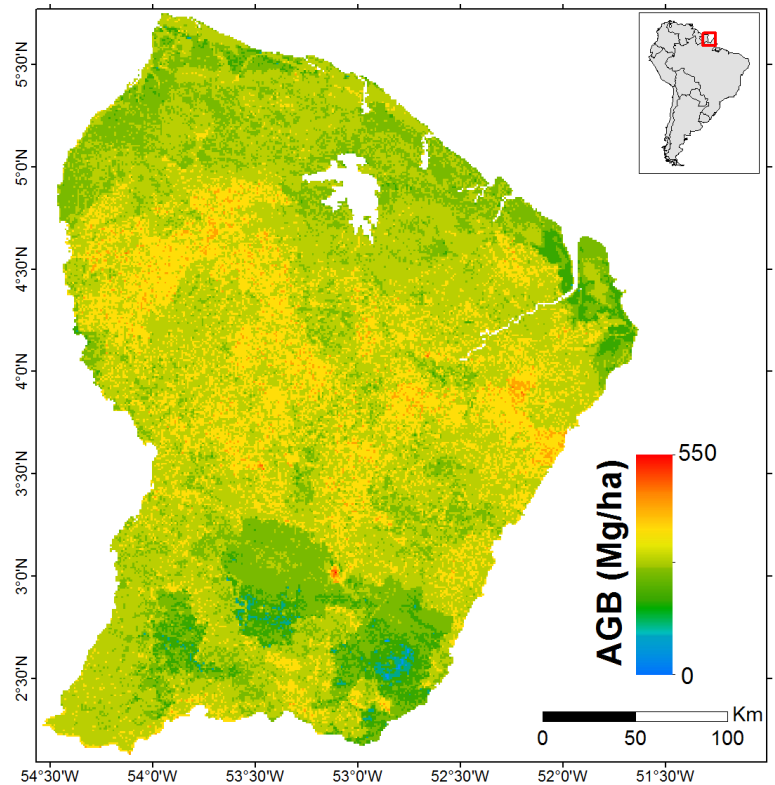

## 4.3. Comparison to global AGB map

In this section the accuracies of the most recent global AGB map produced by Avitabile et al., 2015 are assessed. Avitabile et al., 2015 AGB map integrated two other global AGB maps produced respectively by Saatchi et al., 2011 and Baccini et al., 2012 using a fusion model that was based on the concept by Ge et al., 2014 and further customized to their study. The fusion model consisted of bias removal and the weighted linear averaging of Saatchi's and Baccini's maps to theoretically produce a more precise map than both with decreased bias (Avitabile et al., 2015). The bias removal method consisted of adding the average difference between the input map (Baccini or Saatchi maps) and the reference data for each stratum to the final product.

The map was compared to our AGB *in situ* reference dataset in the same manner used to validate our own product. Results showed that Avitabile's map overestimated AGB by 73.0 Mg/ha compared to the *in situ* AGB dataset (Figure 11). Moreover, the precision of Avitabile's map over French Guiana was in the order of 104 Mg/ha ($R^2$ of 0.06).

**Figure 11. Comparison between our *in situ* AGB estimates and the AGB estimates from Avitabile et al., 2015.**

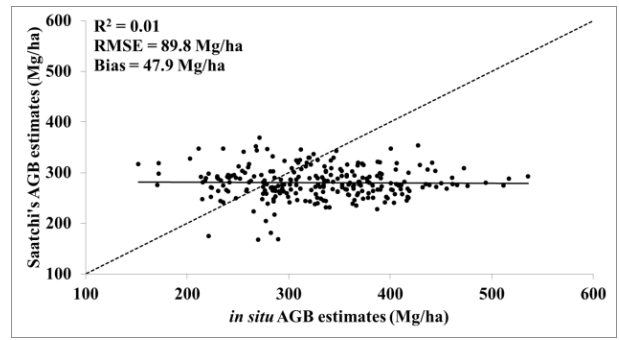

### 4.4. Comparison to the AGB estimates from P-band Radar

The AGB map that was produced in this study was compared to the AGB estimates by Tong Minh et al., 2014 over two small study sites, namely, Paracou (~30 km$^2$) and Nouragues (~17 km$^2$), in French Guiana. Tong Minh et al., 2014 AGB estimates were obtained by tomography with airborne P-band radar data. The AGB estimates map had a resolution of 50 m and when resampled to 1km (by averaging the 50-m pixels in each 1-km pixel), exhibited an RMSE that was better than 30.2 Mg/ha and an $R^2$ of 0.71 compared to our *in situ* AGB estimates. A comparison between our AGB map and the AGB maps from the study of Tong Minh et al., 2014 showed very slight bias (~7 Mg/ha), with a precision for the AGB estimates of 48.7 Mg/ha and an $R^2$ of 0.38 (Figure 12).

**Figure 12. Comparison between our AGB estimates and the AGB estimates from the study of Tong Minh et al., 2014.**

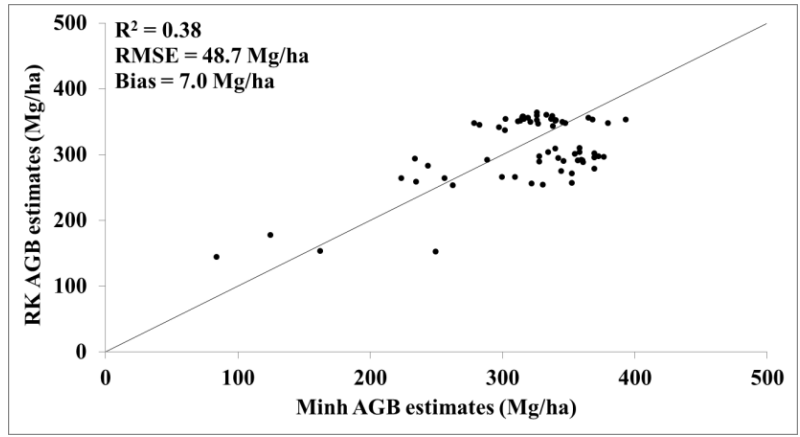

### 4.5. Reported total carbon stock from different AGB maps

In this section, the AGB stock that was produced in different maps was analyzed. The total carbon stock for the 1-km AGB maps was calculated by using the following formula:

$$carbon\ stock\ (t\ C) = \sum_{i=1}^{n}(AGB_i\ in\ Mg/ha) * 0.01 * 0.5 \qquad (6)$$

where n is the total number of available pixels on the AGB map, 0.01 is the conversion factor from hectares to square kilometers, and 0.5 is the conversion ratio from biomass to carbon.

The carbon stocks in French Guiana were estimated to be 1 323 010 kt C according to the AGB map from the regression Kriging process (Table 1). We also estimated the carbon stocks from a recent AGB map over French Guiana ( Guitet et al., 2015). Guitet et al., 2015 used predictive models that considered spatial and environmental effects to predict the AGB from AGB plot estimates and environmental variables. A comparison of the carbon stocks from equation 3 showed that our estimates were within 4% of the estimates from the study of Guitet et al., 2015. However, the carbon stock estimates in this study were 17.6% and 16.2% higher compared to the studies of Baccini et al., 2012 and Saatchi et al., 2011, respectively. Finally, a comparison between our carbon stock estimates and those from Avitabile et al., 2015 showed that our estimates were 34.48% lower (Table 1).

**Table 1. Comparison of the carbon stock estimates in French Guiana (resolution of 1 km) from different studies (the carbon stock is in kilotons of carbon).**

| Study | Carbon stock (kt C) |
|---|---|
| Regression Kriging | 1 323 010 |
| Guitet et al., 2015 | 1 283 320 |
| Baccini et al., 2012 | 1 109 000 |
| Saatchi et al., 2011 | 1 125 000 |
| Avitabile et al., 2015 | 1 874 220 |

## 5. Discussion

In this study, we produced an AGB map of French Guiana with a medium resolution (1 km) by using a mix of spaceborne LiDAR, optical and radar data together with additional environmental datasets. This approach allowed us to capture the large variations in AGB from permanent environmental effects more local contrasts from long-term forest dynamics. The mean absolute percentage error of the AGB estimates at the country level was approximately 12%.

Additionally, a strong correlation between *in situ* AGB estimates and GLAS metrics was present. Although important spatial (250 m distance between the *in situ* AGB estimates and the GLAS footprints) and temporal (*in situ* AGB dataset that was created decades before the GLAS dataset) shifts occurred between the GLAS dataset and the corresponding *in situ* AGB estimates, the precision of our map was better than main carbon benchmarks at this scale (Table 1). To date, researchers have preferred to select a small number of plots to guarantee the quality of the terrain data (inventory with botanical determinations, including trees with small DBH, etc.) to calibrate biomass estimation models, which poses a risk of over-representing the most complex vegetation. We assume that calibrating with data that might be less precise but more abundant is more beneficial, which allows us to better comprehend the regional variations. Therefore, using a maximum number of forest-census field data to calibrate AGB spatial models, not only field plots that exactly matched the locations of remote-sensing footprints, seems to be a good strategy to improve the accuracy of AGB maps. A limitation of the approach in this study is the unavailability of new GLAS measurements since the satellite ceased operations in 2009. However, implementing the approach in this study for different forest sites is possible because of the large global database of GLAS footprints (hundreds of thousands to millions of footprints available across almost every country). Although none of the methodological steps that we developed were entirely novel, the combination of various types of datasets allowed us to produce AGB maps of a dense tropical forest with much more detail than existing maps. Indeed, we obtained similar overall precision than a previous AGB map that was produced over French Guiana with a grid size of 1 km by using only environmental variables without remote-sensing data (Guitet et al., 2015). However, while both quantitative precisions were similar, the qualitative precision of our map appeared to be higher. Indeed, the environmental variables very successfully distinguished large-scale variations at the landscape scale but insufficiently captured AGB variations over short distances because of the natural forest dynamics. Introducing the canopy height component to the model allowed us to better differentiate the local variations from uncommon forest types that were too difficult to predict by using only environmental predictors (e.g., low forests on shallow soils, very open swamp forests among moist forests, etc.). For example, we can distinguish clear AGB variation patterns from terra firma forests to coastal mangroves along the flat eastern coast of the territory because of the intensity of water flooding, which was not as clear on the previous environmental model (Figures 13a & 13b). Similarly, we can clearly observe AGB variations between local patches of white-sand forests and low canopies that are surrounded by highest forests with more biomass along the western coastal plain (Figures 13c & 13d). This approach also allowed us to distinguish larger variations in AGB within the same landscape class from the long-term turnover of forest structure. For example, we can distinguish a clear exposure effect on Mont Itoupe, the highest relief of French Guiana, which has very high biomass on the western slope (downwind), very low biomass on the summit (covered by low-canopy cloud forest) and intermediate biomass on the eastern slope (windward). This result could be obtained by using the canopy height, which adds a variable that represents the vertical characteristics of a forest to the AGB mapping model.

A comparison between the map that was produced in this study and two recent AGB maps, the first being a global AGB map by Avitabile et al., 2015, and the other are AGB estimates over two small sites in French Guiana by Tong Minh et al., 2014 using P-band radar data. The comparison was done against the *in situ* AGB estimates. Results showed very large discrepancies for the global map, while the AGB maps precision from P-band radar were slightly at par with our product. The comparison to the AGB map of Avitabile et al., 2015, which was created using a data fusion of

Baccini and Saatchi's maps, showed poor precision compared to our *in situ* AGB dataset. The RMSE of the AGB estimates was on the order of 100 Mg/ha, mostly because of the low frequency of available calibration points over the dense neo-tropical forests that were used in Avitabile et al., 2015 AGB estimation model. This model, which is calibrated with few reference points, tends to be more sensitive to outliers. Regression models are sensitive to calibration data, so outliers can easily affect the calibration results if few reference points are used. Indeed, this model was mainly calibrated with reference points along the margins of tropical regions (Mexico and Australia) and thus became strongly determined by those reference points outside the most humid areas, which contained the highest stocks of biomass. The bias (our *in situ* AGB estimates - estimates from Avitabile's map) was also significantly high, with Avitabile's estimates being ~73 Mg/ha higher. This result can be explained by Avitabile's technique to reduce the bias of Baccini and Saatchi's maps. The bias that was estimated by computing the average difference between the input map (Baccini or Saatchi's maps) and the reference data for each stratum was added to the final product ( Avitabile et al., 2015). The small number of reference points in Avitabile's study over French Guiana (average AGB of ~420 Mg/ha), caused the AGB to be overestimated by ~73 Mg/a.

Comparison results of AGB from Tong Minh et al., 2014 with our in situ AGB estimates were far better in comparison to the global map, with a precision on the AGB estimates of 30.2 Mg/ha ($R^2$ of 0.71). The low-frequency radar (P-band) is sensitive to the dielectric constant of woody elements, which is linked to the biomass of these elements for a canopy structure Family (dense, sparse, tropical, boreal, etc.). These relationships are less 'variables' as those that bind the height to biomass; we can therefore expect it to work more globally when we have separated the family structures. While P-band data's coverage is low, and costly, in the future, biomass mapping at regional or global scales should be more appropriate with the long wavelength SAR (P-band) than spaceborne LIDAR mapping, since they allow regular mapping without being disturbed by weather conditions. P-band's sensitivity to biomass decreases after ~ 300 t / ha, and is often obscured by topographic effects. Recent studies on topographic correction, taking into account both the geometrical effect and scattering mechanisms ( Villard and Le Toan, 2015) helped usher a new backscattering coefficient in order to minimize the effects of topographic slopes and enhance the sensitivity of the radar signal up to 480 Mg/ha (French Guiana). In addition, with the recent development of tomography, the ground effects are reduced by having access to layers inside the forest canopy where the backscatter from vegetation-ground interactions is not significant ( Tong

Minh et al., 2014). This way, forest biomass can be investigated by considering not only the backscatter intensity at each location but also its vertical distribution. In fact, tomography allows an access up to 10 m in the vegetation cover, where the biomass is well connected to the total aboveground biomass. The signal's sensitivity to the biomass from this layer does not decrease; no signal saturation effect is detected for AGB levels up to 500 Mg/ha (RMSE around 11%) ( Tong Minh et al., 2014). The BIOMASS mission (P-band) launching in 2020 at the earliest will provide multi-temporal global forest AGB maps (ESA 2012). One of the mission objectives is to map the world's forest ASB, for the needs of national scale inventory and global carbon flux calculations. This mission will provide biannual biomass maps of all rainforests and part of the boreal and temperate forests unaffected by the US ground radar ban, with pixels of four ha (200 m x 200 m) and an uncertainty of 20% maximum (ESA 2012).

**Figure 13. Comparison between the AGB estimates from this study when using the regression Kriging technique (a and c) and AGB estimates from the study of Guitet et al., 2015 (b and d) over the eastern coastal plains (a & b) and the western coastal plains (c & d).**

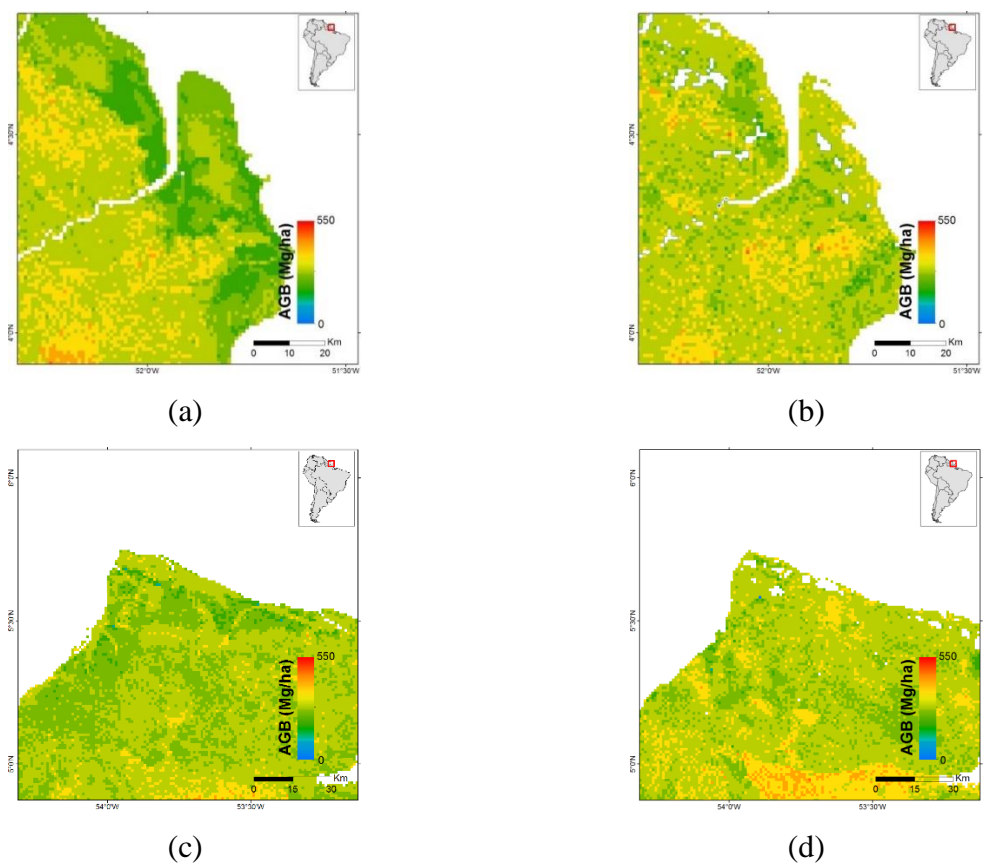

## 6. Conclusions

An approach for AGB mapping over the tropical forest of French Guiana was presented. This approach was based on the merging of *in situ* AGB estimates, GLAS AGB estimates and ancillary data (optical, radar and environmental data). First, the AGB was estimated by using GLAS data and calibrated *in situ* AGB points that were distanced 250 m (at maximum) from the GLAS footprints. The choice of this distance was based on a compromise between the number of calibration points and their correlation to the AGB. Next, a LM was created to estimate the AGB with the most important GLAS metrics. This model showed an RMSE for the estimation of AGB of 48.3 Mg/ha ($R^2$ of 0.54). These results showed that neither a temporal mismatch between the acquisition dates of GLAS and in situ AGB data nor a spatial shift (maximum of 250 m between the GLAS and in situ AGB data) seemed to be a problem for AGB spatial modelling.

We used the regression Kriging technique with the random forests algorithm for the regression portion for the wall-to-wall mapping of AGB. This approach allowed us to achieve a precision for the AGB estimates of 50.2 Mg/ha and an $R^2$ of 0.66 for a 1000-m grid-size map. The precision when using the regression Kriging technique was 72.8 Mg/ha and 43.2 Mg/ha for maps with cell sizes of 500 and 2000 m, respectively.

**Acknowledgments**

The authors wish to thank the French Space Study Center (CNES, DAR 2014 TOSCA) for supporting this research. The authors acknowledge the National Snow and Ice Data Center (NSDIC) for the distribution of the ICESat/GLAS data. The authors also acknowledge the French Geological Survey (BRGM) and, in particular, José Perrin for providing the low-density LiDAR dataset. The authors wish to thank Lilian Blanc (Cirad) and Grégoire Vincent (IRD) for providing the high-density LiDAR dataset. We would also like to thank Noveltis and Airbus Defense and Space for their financial support. We extend our thanks to the French Forest Agency (ONF) and the Guianese National Park (PAG) for providing field data with funds from the French Ministry of the Environment's ECOTROP program and the European Union's PO-FEDER program HABITATS. Finally, we wish to acknowledge that this work has benefited from an

"Investissement d'Avenir" grant that was managed by the Agence Nationale de la Recherche (CEBA, ANR-10-LABX-25-01).